\documentclass[runningheads]{llncs}

\usepackage[T1]{fontenc}
\PassOptionsToPackage{table}{xcolor}
\usepackage{tcolorbox}
\usepackage{booktabs}
\usepackage{multirow}
\usepackage{array}
\usepackage{graphicx,verbatim}
\usepackage{amsmath}
\usepackage{amssymb}
\usepackage{mwe}
\usepackage{hyperref}
\hypersetup{hypertexnames=false}
\usepackage{url}
\usepackage{tabularray}
\usepackage{float}
\usepackage{bbding}
\usepackage[table]{xcolor}
\usepackage[normalem]{ulem}    
\usepackage{enumitem}

\begin{document}

\title{ASTAR: Automated induction of STAndardized radiology Reporting templates from large-scale clinical free-text corpora}
\titlerunning{ASTAR: Automated induction of STAndardized Reporting templates}

\author{
Xinfeng Zhang\inst{1}$^{\dagger}$ \and
Mingxuan Liu\inst{1}$^{\dagger}$ \and
Yifei Chen\inst{1} \and
Juncheng Zhu\inst{2} \and
Kasidit Anmahapong\inst{1} \and
Yiming Huang\inst{3} \and
Yuan Zhang\inst{4} \and
Hongjia Yang\inst{1} \and
Yi Liao\inst{2} \and
Gang Ning\inst{2} \and
Haibo Qu\inst{2} \and
Qiyuan Tian\inst{1}$^{*}$
}

\authorrunning{X. Zhang et al.}

\institute{
Tsinghua University, Beijing, China \and
Sichuan University, Chengdu, China \and
University of California San Diego, San Diego, USA \and
Southeast University, Nanjing, China\\
\email{qiyuantian@tsinghua.edu.cn}
}
\maketitle

\begingroup
\renewcommand{\thefootnote}{}
\footnotetext{$^{\dagger}$ Equal contribution. \quad $^{*}$ Corresponding author.}
\endgroup

\begin{abstract}
Structured reporting converts free-text radiology narratives into queryable data keys, facilitating cohort assembly, longitudinal tracking, and training label generation for medical AI. The prevailing paradigm follows a two-stage pipeline: (1) constructing a reporting template, (2) extracting information to populate it. While the extraction stage has benefited from advances in large language models (LLMs), template construction remains a manual bottleneck relying on labor-intensive expert consensus that is static, difficult to scale, and may fail to capture real-world reporting diversity. We address this limitation with \textbf{\texttt{ASTAR}}, an LLM-based framework for Automated induction of STAndardized radiology Reporting templates from large-scale clinical free-text corpora. Extensive experiments on 4,215 fetal brain MRI reports from multiple centers demonstrate that the \textbf{\texttt{ASTAR}}-induced template surpasses two expert-curated templates across template coverage, information fidelity, diagnostic fidelity, and expert-rated usability, reducing template development from weeks of committee deliberation to hours of automated processing. Code: \url{https://github.com/birthlab/ASTAR}
\end{abstract}

\section{Introduction}
Radiology departments worldwide generate clinical reports at an ever-growing
scale. In the United States alone, over 16.1 million CT examinations were
reported during a nine-month period in 2020~\cite{Davenport2021}, and imaging
utilization across all modalities is projected to rise 17--27\% by
2055~\cite{Christensen2025}. Each examination produces a detailed free-text
narrative synthesizing categorical descriptors, quantitative measurements, and
diagnostic impressions, thereby forming  one of the richest data repositories in modern
healthcare~\cite{nowak_privacy-ensuring_2025,chen2025mmlnbmultimodallearningneuroblastoma}. However, the utility of these reports is constrained because findings remain archived as unstructured text with wide stylistic, terminological, and structural variations across radiologists, institutions, and languages~\cite{liu_fetalextract-llm_nodate,nowak_transformer-based_2023}. Such heterogeneity hinders systematic case retrieval, cohort assembly, longitudinal tracking, and training-label generation for medical AI~\cite{chen_burextract-llama_2024,nowak_privacy-ensuring_2025}. Radiology societies therefore increasingly advocate for structured reporting~\cite{sofia_standardised_2024,busch_large_2024}, converting narratives into queryable keys that support rapid cohort identification and high-quality ground-truth generation~\cite{bai_chest-omdl_nodate,bai2026exactexplainableanomalyawarevision}.

Nevertheless, achieving these benefits at scale requires automated tools, since manual structuring of the massive and continuously growing volume of narrative reports is highly impractical. Therefore, recent studies increasingly employ natural language processing (NLP) and large language models (LLMs) for automated information extraction \cite{fytas-etal-2024-rule,Zhang2026ContentGeneration}. Specifically, the prevailing paradigm follows a two-stage pipeline: (1) a reporting template (i.e., a target template defining the set of keys and their permissible values) is constructed; and (2) relevant information is extracted from free-text reports to populate this template. For example, BURExtract-Llama \cite{chen_burextract-llama_2024} extracts clinical concepts from breast ultrasound reports with an average F1 score of 84.6\%.

\begin{figure}[t]
  \centering
  \includegraphics[width=0.991\textwidth]{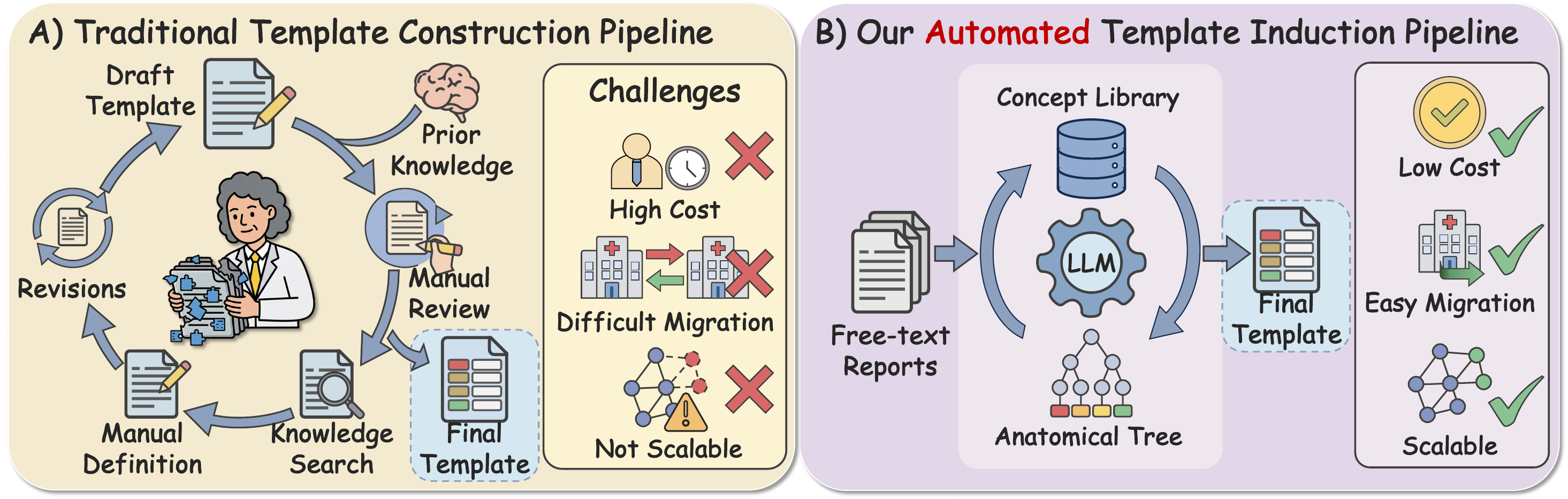}
  \caption{(A) Traditional template construction pipeline. (B) Our automated template induction pipeline (\textbf{\texttt{ASTAR}}).}
  \label{fig:overview}
\end{figure}

However, while the extraction stage has benefited substantially from recent advances in NLP and LLMs, the template construction stage, as an upstream prerequisite for all extraction methods, remains a manual bottleneck, since existing templates are typically produced through labor-intensive expert consensus (Fig. \ref{fig:overview}A). For example, the ESPR template \cite{sofia_standardised_2024} was derived from multidisciplinary panel discussions, and the JSON template used by FetalExtract-LLM \cite{liu_fetalextract-llm_nodate} was compiled from ESPR guidelines \cite{sofia_standardised_2024}, MeSH \cite{lipscomb2000medical}, and a medical dictionary. Beyond this scalability limitation, expert-designed templates are inherently static and may fail to capture the full spectrum of linguistic diversity and evolving clinical descriptions present in massive historical report corpora \cite{busch_large_2024}. As a result, rare but clinically meaningful findings risk being absent from the predefined template, while institution-specific reporting conventions cannot be accommodated by a one-size-fits-all structure.

To the best of our knowledge, no automated method exists for standardized radiology reporting templates construction. To fill this gap, we propose \textbf{\texttt{ASTAR}} (Automated induction of STAndardized radiology Reporting templates), an LLM-based framework (Fig. \ref{fig:overview}B) that automatically mines unified reporting structures from large-scale radiology report corpora. We demonstrate its utility on 4,215 fetal brain MRI reports from three centers, where the \textbf{\texttt{ASTAR}}-induced template achieved stable information retention with progressive consolidation of long-tail keys as corpus size increases, offering a scalable, data-driven complement to labor-intensive expert consensus.

\section{Method}

\subsection{Overview of \texttt{ASTAR}}
\begin{figure}[htbp]
  \centering
  \includegraphics[width=0.99\textwidth]{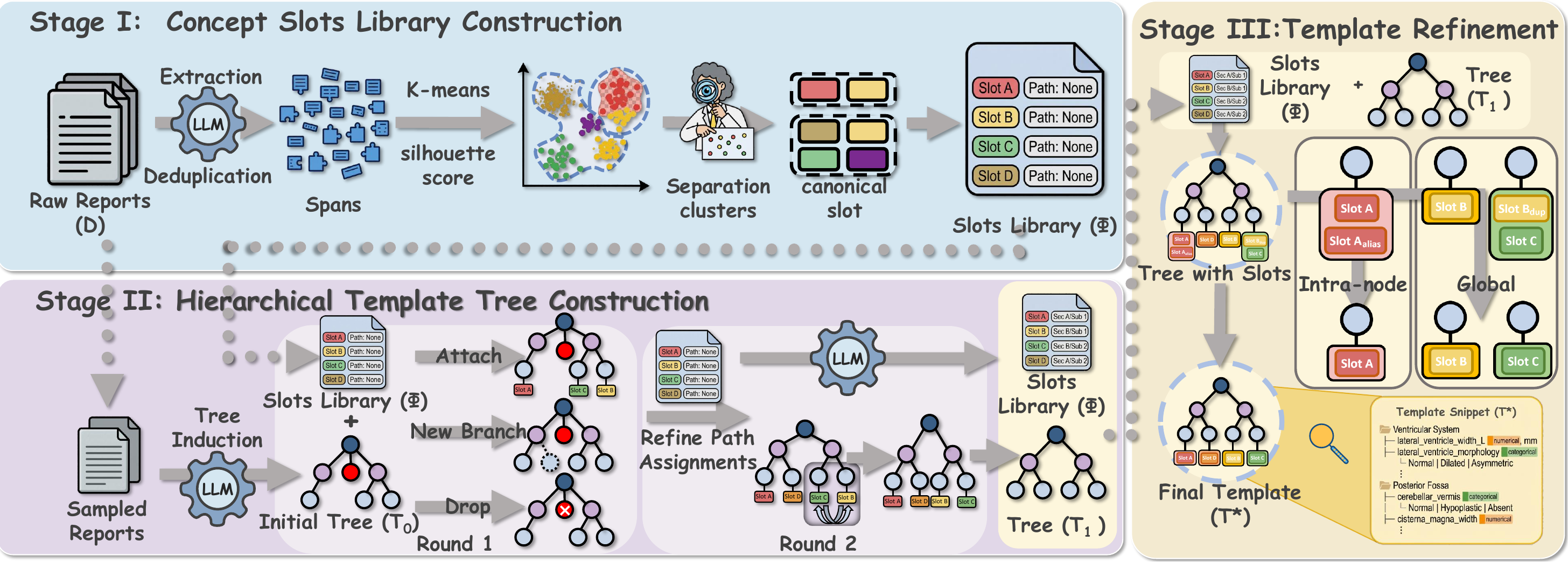}
  \caption{Overview of \textbf{\texttt{ASTAR}}. (1) Stage I: Concept Slots Library Construction. (2) Stage II: Hierarchical Template Tree Construction. (3) Stage III: Template Refinement.}
  \label{fig:pipeline}
\end{figure}

Given a corpus of $N$ de-identified free-text radiology reports
$\mathcal{D}=\{r_1,\dots,r_N\}$, the goal of \textbf{\texttt{ASTAR}}
is to automatically induce a hierarchical reporting template
$\mathcal{T}=(\mathcal{V},\,\mathcal{E},\,\{f_v\}_{v\in
\mathcal{V}_{\text{leaf}}})$, where internal nodes represent anatomical
or semantic groupings and each leaf node
$v\in\mathcal{V}_{\text{leaf}}$ carries a field descriptor
$f_v=(k_v,\tau_v,\Omega_v)$ specifying a standardized field name~$k_v$,
a value type
$\tau_v\in\{\texttt{categorical},\allowbreak\texttt{numerical},\allowbreak\texttt{free\mbox{-}text}\}$,
and a type-dependent value specification~$\Omega_v$.
\textbf{\texttt{ASTAR}} obtains~$\mathcal{T}$ through three stages
(Fig.~\ref{fig:pipeline}):
$\mathcal{D}\xrightarrow{\textbf{I}}\Phi
\xrightarrow{\textbf{II}}T_1
\xrightarrow{\textbf{III}}\mathcal{T}^{*}$,
where Stage~I builds a de-duplicated concept slots library
$\Phi=\{\phi_1,\dots,\phi_L\}$ with each
$\phi_j=(k_j,\tau_j,\Omega_j)$ sharing the same structure as~$f_v$;
Stage~II organizes~$\Phi$ into a hierarchical tree, mapping each
surviving slot to exactly one leaf
($f_v\!\leftarrow\!\phi_j$); and Stage~III refines the tree into the
final template~$\mathcal{T}^{*}$.

\subsection{Stage I: Concept Slots Library Construction}

In stage I, raw narratives are converted into a de-duplicated library of canonical clinical concept slots in two steps. (1) \underline{Span Extraction.} An LLM (Qwen-Max~\cite{qwen3}) in JSON-mode segments each report $r_i$ into spans $\mathcal{S}_i$, each represented as a 7-tuple $(a_{\text{struct}},\,a_{\text{sub}},\,a_{\text{attr}},\, a_{\text{type}},\,a_{\text{val}},\,a_{\text{role}},\, a_{\text{sec}})$ encoding anatomical location (structure / substructure), attribute identity (name / type), a canonicalized value (numerics abstracted as $\langle\texttt{NUM}\rangle$), clinical role ($\texttt{finding}\mid\texttt{measurement}\mid\texttt{diagnosis}$), and report section ($\texttt{description}\mid\texttt{impression}$). Spans from all reports are pooled and de-duplicated by the composite key $(a_{\text{val}},a_{\text{role}})$ into a global span set~$\mathcal{S}$. (2) \underline{Clustering and Slot Induction.} Each span is embedded via Qwen3-Embedding-8B~\cite{qwen3} and grouped by $K$-means into $C$ clusters, with the optimal 
number of clusters $C^{*}$ chosen by maximizing the silhouette 
score~\cite{ROUSSEEUW198753} over $C\!\in\![50,300]$. Because raw clusters may still mix heterogeneous concepts (e.g., \textit{ventricular width} and \textit{ventricular morphology}), the LLM sub-partitions each cluster into semantically coherent subgroups and induces a canonical slot $\phi=(k,\,\tau,\,\Omega)$ per subgroup, where $k$ is a standardized key name, $\tau\!\in\!\{\texttt{categorical},\texttt{numerical}, \texttt{free-text}\}$ the value type, and $\Omega$ the permissible value set. Slots sharing identical standardized key names (k) are then globally merged by the LLM, yielding the unified concept slots library $\Phi=\{\phi_1,\dots,\phi_L\}$.

\subsection{Stage II: Hierarchical Template Tree Construction}

In Stage~II, given the slots library $\Phi$, the flat set of slots 
is organized into a clinically meaningful hierarchy through a 
two-round LLM-driven process.

\noindent\textbf{Round~1: Template Construction and Initial Assignment.}
An LLM (Qwen-Max) first generates an initial Template $\mathcal{T}_0$ from a 
representative subset of reports, with internal nodes organized by 
major anatomical regions. Each slot $\phi_j \in \Phi$ is then 
presented to the LLM together with its descriptive statistics 
and the current template. The LLM decides to \texttt{attach} it 
under an existing node, \texttt{propose} a new branch, or 
\texttt{drop} it. Proposed branches are incorporated into the 
template incrementally.
\noindent\textbf{Round~2: Global Path Refinement.}
Because Round~1 slots are assigned against an evolving template,
early slots never see branches proposed later. A second pass 
therefore re-evaluates all surviving slots in batches on the 
now-complete tree, taking each slot's Round~1 path as reference, 
and outputs a revised path, yielding $\mathcal{T}_1$.

\subsection{Stage III: Template Refinement}

The final stage performs two refinement operations on $\mathcal{T}_1$ to ensure global consistency and output the final template $\mathcal{T}^*$. (1) \underline{Intra-node Semantic Reorganization.} Within each node, the LLM consolidates synonymous permissible values (e.g., [\textit{``normal''}, \textit{``no abnormality''}, \textit{``unremarkable''}] $\to$ [\textit{``Normal''}]) and harmonizes value type assignments across sibling keys, reducing value-level redundancy while preserving clinical granularity. (2) \uline{Global Concept Harmonization.} A tree-wide pass identifies cross-node redundancies, i.e., keys whose names differ across subtrees but refer to the same concept, and standardizes terminology throughout the tree (e.g., unifying \textit{``cerebral hemisphere''} vs.\ \textit{``brain hemisphere''} across subtrees). 

\subsection{Evaluation Protocol for Radiology Reporting Templates}
\label{sec:eval}

Because there is no widely used protocol for evaluating radiology reporting templates, we introduce a three-level protocol to fill this gap: (1) \uline{Template quality} assesses whether the template is comprehensive and well-structured; (2) \uline{structuring fidelity} measures whether populating the template preserves the information in the original report; and (3) \uline{radiologist judgment} captures whether clinicians find the template usable and clinically valuable. Specifically,

(1)~\uline{Template quality}, computed on a held-out test dataset, quantifies how completely $\mathcal{T}^*$ captures the clinical concepts present in unseen reports, measured by \emph{case-level} (macro-averaged over reports) and \emph{key-level} (micro-averaged over all keys) coverage. For each test report $r_i$, concept keys $K_i$ are extracted by concatenating entity and attribute keys of each span. An LLM (Qwen-Max) judges whether each key has a semantic match in the template key set $\mathcal{F (\mathcal{T}^*)}$. The two metrics are defined as:
\begin{equation}
  \mathrm{Cov}_{\text{case}}
  = \frac{1}{|\mathcal{D}_{\text{test}}|}
    \sum_{i}
    \frac{\sum_{k \in K_i} \mathrm{match}(k,\mathcal{F})}{|K_i|},
  \quad
  \mathrm{Cov}_{\text{key}}
  = \frac{\sum_{i}\sum_{k \in K_i} \mathrm{match}(k,\mathcal{F})}
         {\sum_{i}|K_i|}.
  \label{eq:cov}
\end{equation}

(2)~\uline{Structuring fidelity}, also computed on the held-out test dataset, measures whether populating the template preserves the information in the original report, assessed via \emph{information fidelity} and \emph{diagnostic fidelity}. Specifically, (a) \emph{information fidelity}, evaluated through a four-step extract--reconstruct round trip, quantifies textual information preservation. \textit{(i)}~An LLM (Gemini-2.5-Pro-Preview~\cite{comanici2025gemini}),
chosen independently of the template-construction LLM to prevent extraction bias, populates the template from each
report's \texttt{imaging description}; \textit{(ii)}~extracted key--value pairs are mapped into the full hierarchy and re-flattened for cross-template consistency; \textit{(iii)}~another LLM (Qwen-Max) reconstructs a free-text report solely from the structured report, guided by a fixed style-reference text; \textit{(iv)}~the reconstruction $\hat{r}_i$ is compared with the original $r_i$ using ROUGE-1/2/L~(F1)~\cite{lin-2004-rouge}, chrF~\cite{popovic2015chrf}, BERTScore$_R$~(F1) (with RoBERTa-wwm-ext~\cite{cui2021pre}), and BERTScore$_M$~(F1)~\cite{zhang2020bertscore} (with MacBERT~\cite{cui2020revisiting}). (b) \emph{Diagnostic Fidelity}, evaluated through an infer-then-judge protocol, tests whether the structured report  retains sufficient clinical information for diagnosis. Qwen-Max receives only the structured output, infers a diagnosis $d_i^{\text{struct}}$ under strict information closure, and an LLM-as-Judge protocol (with Qwen-Max) is used to compare it against the ground-truth impression $d_i^{\text{gt}}$ (i.e., the \texttt{diagnostic impression} in the original free-text report) using three metrics (all normalized to $[0,1]$):
\emph{Primary Diagnosis Accuracy}
  (PDA; 0--5 integer, macro correctness),
\emph{Key Finding Preservation}
  (KFP; 0--1 continuous, abnormal-finding retention),
and \emph{Clinical Actionability}
  (CA; 0--5 integer, decision-support sufficiency).

(3)~\uline{Radiologist judgment} directly evaluates the template
via structured clinician ratings. Two radiologists from different countries, ensuring cross-national diversity in clinical conventions and mitigating institution-specific bias, rated all three templates on 28 Likert-scale items (1--5) across three dimensions. (a)~\emph{Template Structure} (12~items) evaluates the schema design across six sub-dimensions: coverage \& completeness (\textsc{comp}, \textsc{abno}), structure \& granularity (\textsc{logi}, \textsc{gran}) terminology \& clarity (\textsc{name}, \textsc{enum}), safety \& error prevention (\textsc{noms}, \textsc{nofn}, \textsc{unce}), adaptability (\textsc{adap}), and deployment readiness (\textsc{adop}, \textsc{sign}); (b)~\emph{System Usability} (10~items) follows the standard
scale~\cite{brooke1996sus}, comprising
\textsc{freq} (willingness to use),
\textsc{comp} (complexity$^\dagger$),
\textsc{ease} (ease of use),
\textsc{tech} (technical support needed$^\dagger$),
\textsc{inte} (integration),
\textsc{inco} (inconsistency$^\dagger$),
\textsc{lear} (learnability),
\textsc{cumb} (cumbersomeness$^\dagger$),
\textsc{conf} (confidence), and
\textsc{prel} (prior learning needed$^\dagger$),
where $\dagger$ marks negatively worded items reverse-scored
as $6{-}x_{\text{raw}}$; (c)~\emph{Clinical Impact} (6~items) assesses downstream utility: \textsc{noms} (omission reduction), \textsc{cons} (consistency), \textsc{orde} (workflow alignment), \textsc{naex} (not-assessed expressibility), \textsc{comm} (communication efficiency), and \textsc{ovr} (overall satisfaction).


\begin{table}[htbp]
\centering
\caption{Template quality (coverage) and structuring fidelity ((information fidelity \& diagnostic fidelity)) across
templates evaluated on ID ($n{=}100$) and OoD ($n{=}115$)
test sets. Bold indicates best results. The three templates differ in key
count (ESPR: 159, FetalExtract: 51, \texttt{ASTAR}: 88), and the
results suggest that key relevance and organization, rather
than key count alone, contribute to the observed differences.}\label{tab:objective}
\fontsize{8pt}{9.6pt}\selectfont          
\setlength{\tabcolsep}{7pt}               
\renewcommand{\arraystretch}{0.92}         
\begin{tabular}{llccc}
\toprule
Category & Metric & ESPR \cite{sofia_standardised_2024} & FetalExtract \cite{liu_fetalextract-llm_nodate} & \textbf{\texttt{ASTAR}} \\
\specialrule{1.0pt}{1.5pt}{1.5pt}
\textbf{Structure}
 & Number of keys & 159 & 51 & \textbf{88} \\
\specialrule{1.0pt}{1.5pt}{1.5pt}
\rowcolor{black!10}\multicolumn{5}{l}{\textbf{ID test dataset ($n$=100)}} \\
\addlinespace[1pt]
\multirow{2}{*}{\textbf{Coverage}}
 & Case-level & 45.41\% & 36.34\% & \textbf{81.34\%} \\
 & Key-level & 45.76\% & 37.82\% & \textbf{82.42\%} \\
\cmidrule(lr){1-5}
\multirow{6}{*}{\shortstack[l]{\textbf{Information}\\\textbf{Fidelity} }}
 & ROUGE-1 (F1) & 0.651$\pm$0.139 & 0.409$\pm$0.107 & \textbf{0.749$\pm$0.079} \\
 & ROUGE-2 (F1) & 0.513$\pm$0.167 & 0.275$\pm$0.102 & \textbf{0.689$\pm$0.104} \\
 & ROUGE-L (F1) & 0.562$\pm$0.124 & 0.397$\pm$0.103 & \textbf{0.694$\pm$0.100} \\
 & chrF & 0.576$\pm$0.076 & 0.539$\pm$0.093 & \textbf{0.578$\pm$0.075} \\
 & BERTScore$_M$ (F1) & 0.873$\pm$0.016 & 0.865$\pm$0.020 & \textbf{0.877$\pm$0.017} \\
 & BERTScore$_R$ (F1) & 0.892$\pm$0.016 & 0.882$\pm$0.021 & \textbf{0.895$\pm$0.016} \\
\cmidrule(lr){1-5}
\multirow{3}{*}{\shortstack[l]{\textbf{Diagnostic}\\\textbf{Fidelity}  }}
 & PDA & 0.879$\pm$0.119 & 0.858$\pm$0.158 & \textbf{0.953$\pm$0.090} \\
 & KFP & 0.882$\pm$0.134 & 0.844$\pm$0.226 & \textbf{0.958$\pm$0.093} \\
 & CA & 0.876$\pm$0.104 & 0.858$\pm$0.147 & \textbf{0.953$\pm$0.085} \\
\specialrule{1.0pt}{1.5pt}{1.5pt}
\rowcolor{black!10}\multicolumn{5}{l}{\textbf{OoD test dataset ($n$=115)}} \\
\addlinespace[1pt]
\multirow{2}{*}{\textbf{Coverage}}
 & Case-level & 54.32\% & 37.22\% & \textbf{66.12\%} \\
 & Key-level & 55.48\% & 37.09\% & \textbf{66.40\%} \\
\cmidrule(lr){1-5}
\multirow{6}{*}{\shortstack[l]{\textbf{Information}\\\textbf{Fidelity} }}
 & ROUGE-1 (F1) & 0.482$\pm$0.228 & 0.316$\pm$0.215 & \textbf{0.717$\pm$0.264} \\
 & ROUGE-2 (F1) & 0.347$\pm$0.233 & 0.219$\pm$0.227 & \textbf{0.590$\pm$0.318} \\
 & ROUGE-L (F1) & 0.394$\pm$0.175 & 0.305$\pm$0.219 & \textbf{0.503$\pm$0.192} \\
 & chrF & 0.322$\pm$0.094 & 0.269$\pm$0.123 & \textbf{0.376$\pm$0.073} \\
 & BERTScore$_M$ (F1) & 0.843$\pm$0.024 & 0.843$\pm$0.031 & \textbf{0.850$\pm$0.020} \\
 & BERTScore$_R$ (F1) & 0.853$\pm$0.029 & 0.851$\pm$0.033 & \textbf{0.862$\pm$0.023} \\
\cmidrule(lr){1-5}
\multirow{3}{*}{\shortstack[l]{\textbf{Diagnostic}\\\textbf{Fidelity}  }}
 & PDA & 0.790$\pm$0.189 & 0.802$\pm$0.143 & \textbf{0.847$\pm$0.125} \\
 & KFP & 0.733$\pm$0.206 & 0.739$\pm$0.182 & \textbf{0.811$\pm$0.154} \\
 & CA & 0.751$\pm$0.137 & 0.757$\pm$0.134 & \textbf{0.847$\pm$0.107} \\
\bottomrule
\end{tabular}
\end{table}

\begin{figure}[htbp]
  \centering
  \includegraphics[width=\textwidth]{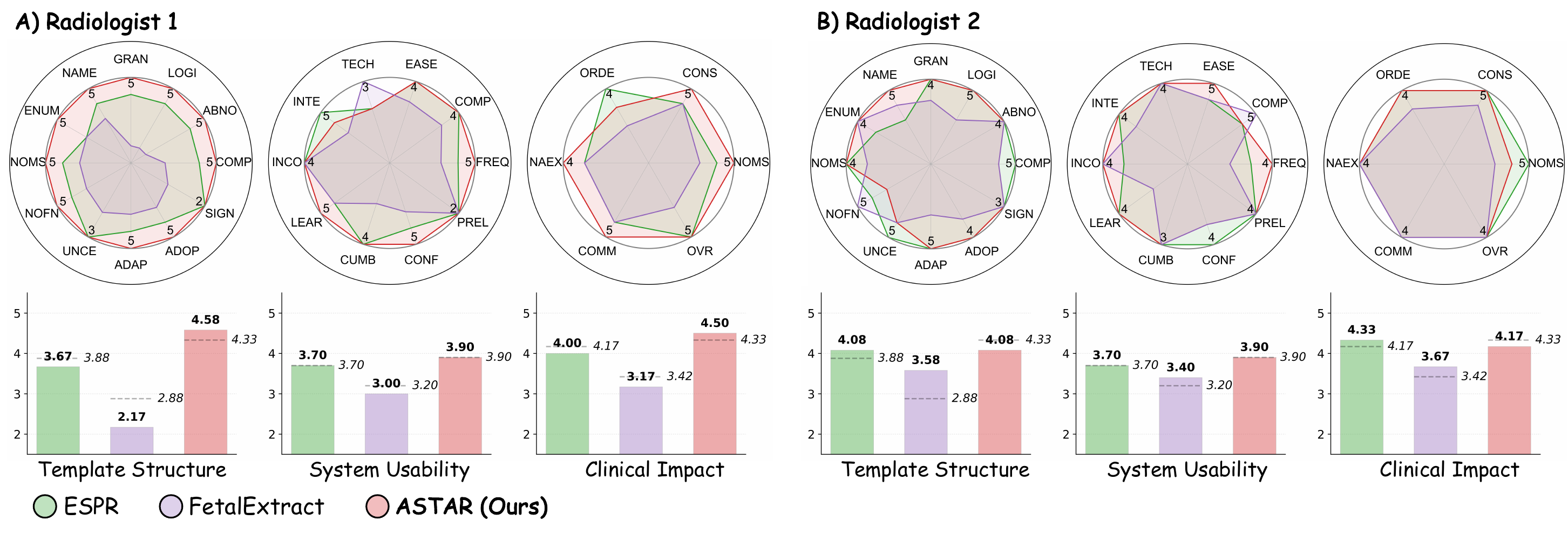}
  \caption{Two radiologists from different countries rated all
templates on 28 Likert-scale items ($1$--$5$, higher is better).
\textbf{(A)} Radiologist 1; \textbf{(B)} Radiologist 2.
\textbf{Top row:} per-item radar charts for Template Structure
(12 items), System Usability (10 items), and Clinical Impact
(6 items); numbers on each axis denote the maximum score among
the three templates for that item. \textbf{Bottom row:}
dimension-level means; dashed lines show cross-radiologist
averages.}
  \label{fig:radar}
\end{figure}

\section{Experiments and Results}
\label{sec:exp}

\subsection{Data Collection}
\label{sec:data}

A total of 4{,}100 de-identified fetal brain MRI
free-text reports were retrospectively collected from the [Anonymous] hospital, covering examinations performed between October 2015 and December 2024 on one 3.0\,T system (Siemens MAGNETOM Skyra) and two 1.5\,T systems (Philips Achieva and United Imaging uMR\,570). Each report comprises a narrative \textit{imaging description} section and a \textit{diagnostic impression} section. Of these, 4{,}000 reports were used for template construction with \texttt{ASTAR} and 100 were held out as an in-distribution (ID) test dataset. To assess cross-institutional generalizability, an additional
out-of-distribution (OoD) test dataset of 115 de-identified
fetal brain MRI reports was collected from [Anonymous] hospital and [Anonymous] hospital.

\subsection{Comparison with Expert-Designed Templates}
\label{sec:baselines}

The ASTAR-induced template was compared against two
expert-designed templates. (1) \uline{ESPR Template}~\cite{sofia_standardised_2024}: a standardized fetal MRI reporting template released in 2024 by the Fetal Task Force of the European Society of Paediatric Radiology. (2) \uline{FetalExtract Template}~\cite{liu_fetalextract-llm_nodate}: a JSON template compiled from ESPR guidelines, MeSH, and a medical dictionary, used for structured information extraction with FetalExtract-LLM.

Across all metrics, \textbf{\texttt{ASTAR}} consistently outperforms both expert-designed templates on both ID and OoD datasets, achieving higher coverage with fewer keys, superior information and diagnostic fidelity, and the highest expert-rated usability (Table~\ref{tab:objective}, Fig.~\ref{fig:radar}). (1)~\uline{Template Quality:} On the ID test dataset, \textbf{\texttt{ASTAR}} achieves case-/key-level coverage of 81.34\%/82.42\%, outperforming ESPR and FetalExtract. This advantage persists on the OoD set, though the gap narrows due to expected domain shift. (2)~\uline{Structuring Fidelity:} \textbf{\texttt{ASTAR}} consistently achieves the highest scores across all six metrics for \emph{information fidelity} on both datasets. Specifically, on the ID dataset, \textbf{\texttt{ASTAR}} attained ROUGE-L $0.694 \pm 0.100$, substantially outperforming ESPR and FetalExtract; BERTScore$_\mathrm{R}$ reached $0.895 \pm 0.016$ vs.\ $0.892 \pm 0.016$ and $0.882 \pm 0.021$. On the OoD test dataset, \textbf{\texttt{ASTAR}} maintains a clear lead (ROUGE-L $0.503 \pm 0.192$ vs.\ $0.394 \pm 0.175$ and $0.305 \pm 0.219$), with larger standard deviations reflecting greater cross-institutional stylistic heterogeneity. For \emph{diagnostic fidelity}, \textbf{\texttt{ASTAR}} achieves the highest scores across all three dimensions on the ID set: PDA $0.953 \pm 0.090$, KFP $0.958 \pm 0.093$, and CA $0.953 \pm 0.085$, outperforming ESPR (PDA/KFP/CA: $0.879$/$0.882$/$0.876$)
and FetalExtract ($0.858$/$0.844$/$0.858$). On the OoD set, \textbf{\texttt{ASTAR}} retains its advantage
(PDA $0.847$, KFP $0.811$, CA $0.847$),
with consistently high KFP ($>$\,0.81) confirming that
the automatically induced template preserves critical
abnormal findings (e.g., ventriculomegaly, posterior fossa
anomalies) essential for clinical decision-making. (3)~\uline{Radiologist Judgment:} \textbf{\texttt{ASTAR}} receives the highest scores across \emph{Template Structure} 4.33/5.00 (vs.\ ESPR 3.88, FetalExtract 2.88), \emph{System Usability} 3.90 (vs.\ 3.70, 3.20), and \emph{Clinical Impact} 4.33 (vs.\ 4.17, 3.42). 

\begin{figure}[t]
  \centering
  \includegraphics[width=\linewidth]{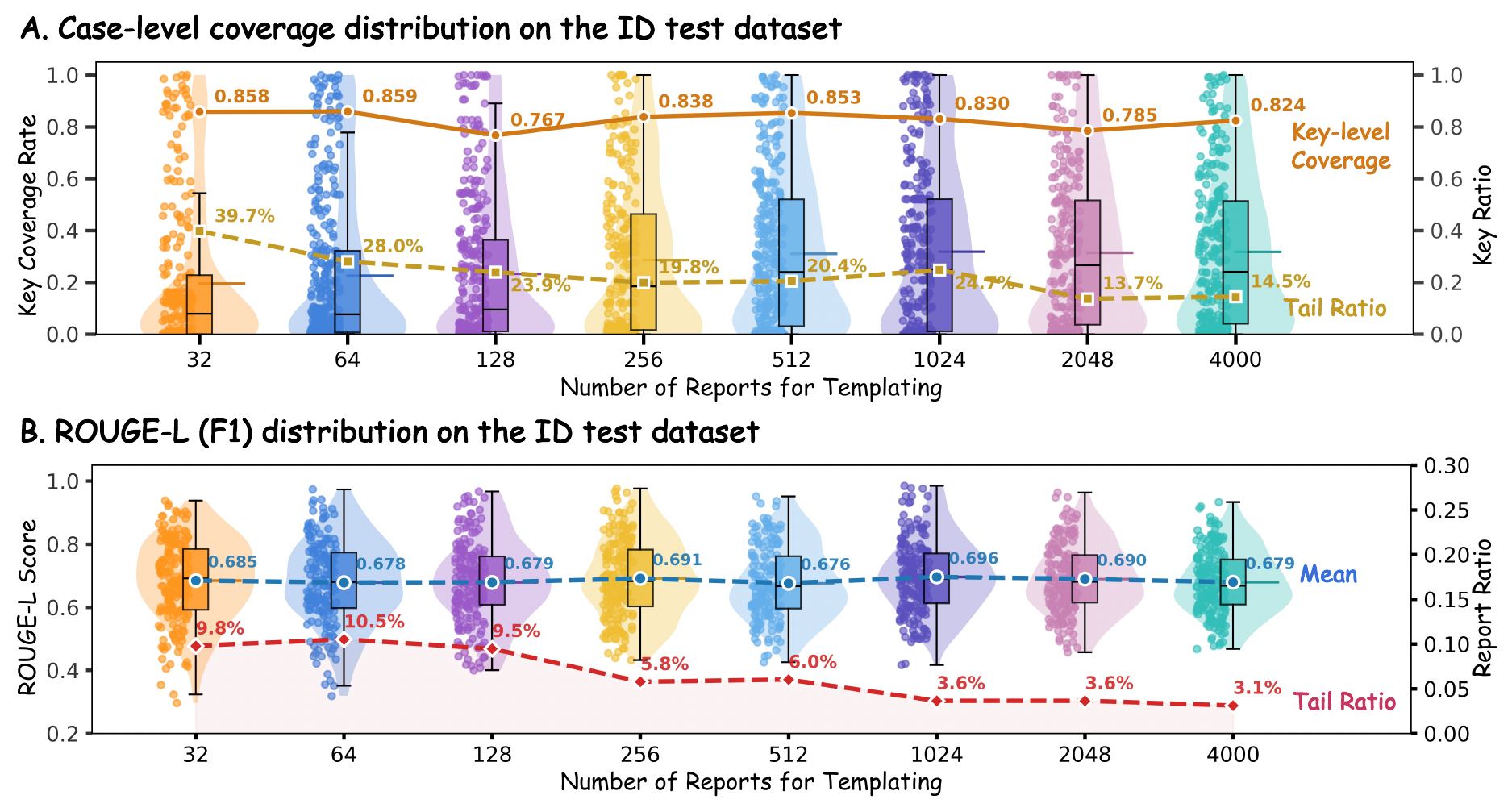}
  \caption{Scaling behavior of ASTAR-induced templates on
  the ID test set ($n{=}100$).
  \textbf{(A)}~Case-level coverage (left axis), key-level coverage (orange, left axis) and
  key coverage tail ratio (gold, right axis).
  \textbf{(B)}~ROUGE-L distributions (left axis), mean ROUGE-L (blue, left axis) and
  report-level tail ratio (red dashed,
  right axis).}
  \label{fig:scaling}
\end{figure}

\subsection{Scaling Behavior of \textbf{\texttt{ASTAR}}}
\label{sec:scaling}

To investigate how corpus size affects template quality, we constructed \textbf{\texttt{ASTAR}} templates from subsets of $n{=}32$ to $n{=}4000$ reports in the training dataset and evaluated each on the ID test dataset (Fig.~\ref{fig:scaling}). Key-level coverage remains consistently high (0.767--0.859), while the tail ratio (case-level coverage < 0.01) decreases from 39.7\% at $n{=}32$ to 14.5\% at $n{=}2048$ before stabilizing, confirming that \textbf{\texttt{ASTAR}} progressively consolidates rare-but-valid concepts into coherent keys (Fig.~\ref{fig:scaling}A). Additionally, mean ROUGE-L remains stable across all corpus sizes  (0.676--0.696), while the report-level tail ratio (case-level ROUGE-L < 0.5) drops from 9.8--10.5\% at $n \leq 64$ to 3.1\% at $n{=}4000$, indicating that additional reports primarily improve coverage of structurally atypical or linguistically rare cases rather than boosting average fidelity (Fig.~\ref{fig:scaling}B).

\section{Conclusion}
\label{sec:conclusion}

We presented \textbf{\texttt{ASTAR}}, the first LLM-based framework that automatically induces standardized radiology reporting templates from large-scale free-text corpora. Evaluated on 4,215 multi-center fetal brain MRI reports, the ASTAR-induced template outperforms two expert-designed baseline templates in template quality, structuring fidelity, and radiologist-rated usability, reducing template development from weeks of expert consensus to hours of automated processing.

\bibliographystyle{splncs04}
\bibliography{mybibliography}

\end{document}